\documentclass[runningheads]{llncs}
\usepackage{amsfonts}
\usepackage{amsmath}
\usepackage{amssymb}
\usepackage{mathrsfs}
\usepackage{xcolor}
\usepackage{graphicx}

\def\det{\mathrm{det}}
\def\nondet{\mathrm{rnd}}
\newcommand{\term}{\textbf}

\title{Representation and Invariance in Reinforcement Learning}
\titlerunning{Representation and Invariance in RL}

\authorrunning{S.\ A.\ Alexander, A.\ P.\ Pedersen}
\author{Samuel Allen Alexander\inst{1} \orcidID{0000-0002-7930-110X}\\
Arthur Paul Pedersen\inst{2} \orcidID{0000-0002-2164-6404}}

\institute{$^{1}$Independent Researcher,
\email{samuelallenalexander@gmail.com}\\
$^{2}$The City University of New York}

\begin{document}

\maketitle

\begin{abstract}
    Researchers have formalized reinforcement learning (RL)
    in different ways. If an agent in one RL framework
    is to run within another RL framework's environments,
    the agent must first be converted, or mapped, into
    that other framework. In this
    paper, we lay foundations for studying
    relative-intelligence-preserving mappability between RL
    frameworks. We introduce a criterion which is
    sufficient for relative intelligence to be preserved
    according to one particular method of measuring
    intelligence. We show that this criterion cannot be
    met when mapping between certain deterministic and
    stochastic RL frameworks, suggesting inherent
    fundamental differences between these different versions of
    RL.
\end{abstract}

\section{Introduction}

\emph{If we changed the rules, would the wise become fools?}
In reinforcement learning (RL), agents and environments interact. The agent's objective is to learn to act in its environment
in order to maximize its rewards.
When an agent interacts with an environment, the agent and the environment
take turns. On the agent's turn, the agent chooses an \term{action} (or
a probability distribution over a set of actions, according to which an
action is randomly selected). The action so chosen is thereupon transmitted
to agent and environment. On the environment's turn, the environment chooses a
\term{percept} to send to the agent in response (or a probability distribution
over a set of percepts, according to which a percept is randomly
selected). The percept so chosen is likewise transmitted to agent and
environment. Each percept includes a numerical \term{reward} and an
\term{observation}. 

This is all simple enough, but there are many different ways to formally
represent RL. These different representations can be organized according to
answers to key questions, such as who goes first, what actions are permitted,
what observations are allowed, and how numerical rewards are issued. Implicit in
treatments of RL is that answers to these questions are inconsequential.  The problem addressed in this paper is whether this is
really so. If answers to such questions are inconsequential to problems for
reinforcement learning, then evaluation of agent performance --- measures of
their relative intelligence --- would be expected to be invariant with respect to
transformations between different RL frameworks.

The present paper develops techniques for understanding the extent to which
scales for measuring agent intelligence are invariant to different RL
representations. To be clear, this is a very complicated subject, and the
present paper should be considered to be an initial step in a thousand-mile
journey. Even the problem of merely measuring RL agent intelligence
in some fixed RL framework is quite difficult. How much more difficult is the
problem of testing preservation of relative intelligence when agents are
converted from one RL framework to another?  This is important because
researchers have been treating RL as if the details are irrelevant, speaking of RL as if there is some core approach in common
that everyone agrees on, whereas in reality this is far from the case.

As a simple illustrative example, if someone proposed an RL framework where
only the reward $0$ were allowed, clearly that framework would be weaker than
the frameworks used in practice. Yet it is not so clear whether the same would
hold if the proposed RL framework allowed only the rewards $\{0,1\}$.
What if it allowed only the rewards $\{-1,0,1\}$? What if it allowed only
natural number rewards? What if it allowed only rational number rewards? What if it
allowed arbitrary real-valued rewards? Would all these frameworks be equivalent,
even though none are equivalent to the $0$-only-reward framework? What does this
question even mean, and how would one even begin to answer it?

The above questions, and others like them, are the
high-level considerations which motivated this paper. But we found the problem
so overwhelmingly complicated that, after exploring the topic for years,
we finally decided to limit this initial paper
to a modest first stab at it. In this paper we will consider only four
specific concrete RL frameworks. These four frameworks are mutually identical
except for two parameters: whether agents are deterministic
or stochastic; and whether environments are deterministic or
stochastic. The question of which of these frameworks are
equivalent to which, is already important, because all four types of
RL are routinely used in practice (often with deterministic agents or environments
masquerading as stochastic through the use of pseudo-random number
generators), whereas a majority of the theoretical literature assumes
both agent and environment are stochastic. It would therefore be
scandalous if these four frameworks were not all equivalent. As a matter
of fact, our analysis suggests they might be significantly non-equivalent.

We will introduce a notion of transformation from one RL framework to another,
and we will show that, if relative intelligence is compared as suggested in
\cite{alexander2019intelligence}, then such
transformations preserve relative intelligence of agents.
Thus, at least in some sense, the existence of such a transformation is a
sufficient condition for one RL framework to be reducible to another---and if
two RL frameworks are mutually reducible to each other, they are in that
sense equivalent to each other.

\section{Preliminaries}

The following definition attempts to explicitly recognise a decision
implicit in much of the RL literature.

\begin{definition}
\label{rlframeworkdefn}
    By a \term{reinforcement learning framework} (or \term{RL framework})
    we mean a triple $(A,E,V)$ where:
    \begin{enumerate}
        \item $A$ is a set whose members are called \term{agents};
        \item $E$ is a set whose members are called \term{environments};
        \item $V:A\times E\to\mathbb R$ is a function assigning to every agent
        $\pi\in A$ and environment $\mu\in E$ a \term{total expected reward}
        $V^\pi_\mu\in\mathbb R$ representing how well $\pi$ performs in $\mu$.
    \end{enumerate}
\end{definition}

While Definition \ref{rlframeworkdefn} is not intended to be as general as possible, it encompasses  standard variants of RL\footnote{Subsumed variants include, for example, those  in which
(i) agents are deterministic while environments need not be
    (as in \cite{hutter2004universal} or \cite{silver2021reward}), (ii) neither agent nor environment need be deterministic
    (as in \cite{legg2007universal}),
    (iii) rewards are multiplied by discount factors, as in
    \cite{sutton2018reinforcement},
    (iv) each percept also
    includes a true-or-false flag indicating whether or not the percept signals the
    start of a new ``episode'' (as in \cite{openaigym}
    or \cite{sutton2018reinforcement}, (v) where rewards are restricted, e.g.\ to $\mathbb Q$
    or (as in \cite{legg2007universal}) some finite subset of $\mathbb Q$, (vi) where available actions vary from turn to turn
    (as in \cite{sutton2018reinforcement} or \cite{rlgames}), (vii) where available actions vary from
    environment to environment
    (as in \cite{openaigym}), (viii) where environments are Markov decision processes (as in most of
    \cite{sutton2018reinforcement}), (ix) where the environment can secretly simulate
        the agent \cite{extendedenvironmentspaper,newcomblike,alexanderpedersen}, (x) where, environments and/or agents must be computable.
}.

What does it mean for one RL framework to be reducible to another?
Imagine you have access to agents which a laboratory designed for RL
framework $\mathscr F$, but the environments you are interested in were designed
for RL framework $\mathscr F'$. The agents designed for framework
$\mathscr F$ can only run in
environments designed for framework $\mathscr F$, so you cannot directly use them.
You must somehow convert them to run in framework $\mathscr F'$. So for every agent
$\pi$ designed for framework $\mathscr F$, you need to transform it into an agent
$\pi^*$ designed for framework $\mathscr F'$.
This transformation should be faithful
in some sense, but what does that mean? This question is vague, but we
can at least say one thing: the transformation should preserve relative
performance. If $\pi$ is better than $\rho$ in framework $\mathscr F$, then $\pi^*$
should be better than $\rho^*$ in framework $\mathscr F'$. But $\pi^*$ and $\rho^*$
perform in framework $\mathscr F'$ environments, whereas $\pi$ and $\rho$ perform in
framework $\mathscr F$ environments. And in this thought experiment, the environments
you care about are in framework $\mathscr F'$. So for every such environment
$\mu$ in framework $\mathscr F'$, the relative performance of $\pi^*$ and $\rho^*$
in $\mu$ should be compared with the relative performance of $\pi$ and $\rho$,
not in $\mu$ itself, as they are incompatible with $\mu$, but rather with some
appropriate transformation $\mu_*$ of $\mu$, where $\mu_*$ is an environment
in framework $\mathscr F$. This motivates the following definition.

\begin{definition}
\label{translationdefn}
    Suppose $\mathscr F=(A,E,V)$ and
    $\overline{\mathscr F}=(\overline A,\overline E,\overline V)$ are RL frameworks.
    A \term{transformation} from $\mathscr F$ to $\overline{\mathscr F}$ is a pair
    $(\bullet^*:A\to \overline A,\bullet_*:\overline E\to E)$
    of functions such that:
    \begin{enumerate}
        \item (Faithfulness) For all $\pi,\rho\in A$ and $\mu\in \overline E$,
            $\overline V^{\pi^*}_\mu<\overline V^{\rho^*}_\mu$
            iff $V^\pi_{\mu_*}<V^\rho_{\mu_*}$.
        \item (Nontriviality 1) There exist $\pi,\rho\in A$,
            $\mu\in \overline E$ such that
            $\overline V^{\pi^*}_\mu<\overline V^{\rho^*}_\mu$.
        \item (Nontriviality 2) There exist $\pi\in A$,
            $\mu,\nu\in \overline E$ such that
            $\overline V^{\pi^*}_\mu<\overline V^{\pi^*}_\nu$.
    \end{enumerate}
\end{definition}

\begin{example}
    Suppose $\mathscr F$ and $\overline{\mathscr F}$
    are two RL frameworks identical in
    every way except that $\mathscr F$ permits rewards from $\mathbb Z$
    but $\overline{\mathscr F}$ only permits rewards from
    $2\mathbb Z$, the set of
    \emph{even} integers. We believe anyone familiar with RL would
    informally consider these two RL frameworks to be equivalent.
    The obvious transformation from $\mathscr F$ to $\overline{\mathscr F}$ is the
    pair $(\bullet^*,\bullet_*)$ defined as follows. For any agent $\pi$
    of $\mathscr F$, let $\pi^*$ be the agent of $\overline{\mathscr F}$
    which results from wrapping
    $\pi$ with an intermediary function that divides all rewards by $2$.
    And for any environment $\mu$ of $\overline{\mathscr F}$, let $\mu_*$ be the
    environment of $\mathscr F$ which takes an input,
    multiplies all the rewards in that
    input by $2$, passes the mutated input to $\mu$, and returns $\mu$'s
    output but with reward divided by $2$.
    Similarly, a transformation from $\overline{\mathscr F}$ to $\mathscr F$ can be
    obtained by replacing $2$ with $\frac12$ above.
\end{example}

\begin{lemma}[Composability]
\label{transitivetranslationlemma}

    Suppose $\mathscr F,\overline{\mathscr F},\overline{\overline{\mathscr F}}$
    are RL frameworks.
    If there is a transformation from $\mathscr F$ to $\overline{\mathscr F}$ and
    a transformation from $\overline{\mathscr F}$
    to $\overline{\overline{\mathscr F}}$, then there is
    a transformation from $\mathscr F$ to $\overline{\overline{\mathscr F}}$.
\end{lemma}

\begin{proof}
    Write $\mathscr F=(A,E,V)$,
    $\overline{\mathscr F}=(\overline A,\overline E,\overline V)$,
    $\overline{\overline{\mathscr F}}
    =(\overline{\overline A},\overline{\overline E},\overline{\overline V})$.
    Assume $(\bullet^*,\bullet_*)$ is a transformation from $\mathscr F$
    to $\overline{\mathscr F}$,
    so $\bullet^*:A\to \overline A$ and $\bullet_*:\overline E\to E$.
    Assume $(\bullet^\dag,\bullet_\dag)$ is a transformation from
    $\overline{\mathscr F}$ to $\overline{\overline{\mathscr F}}$,
    so $\bullet^\dag:\overline A\to \overline{\overline A}$
    and $\bullet_\dag:\overline{\overline E}\to \overline E$.
    Define $\bullet^\ddag:A\to \overline{\overline A}$
    by $\pi^\ddag=(\pi^*)^\dag$
    and define $\bullet_\ddag:\overline{\overline E}\to E$ by
    $\mu_\ddag=(\mu_\dag)_*$.
    It is straightforward to show that $(\bullet^\ddag,\bullet_\ddag)$
    is a transformation from $\mathscr F$ to
    $\overline{\overline{\mathscr F}}$.
\end{proof}

\begin{lemma}
    (Self-reducibility)
    Suppose $\mathscr F$ is an RL framework.
    If $\mathscr F$ is nontrivial, in the sense that $\mathscr F$ contains
    agents $\pi,\rho,\sigma$ and environments $\mu,\nu,\tau$ such that
    $V^\pi_\mu<V^\rho_\mu$ and $V^\sigma_\nu<V^\sigma_\tau$, then
    there is a translation from $\mathscr F$ to itself.
\end{lemma}

\begin{proof}
    Write $\mathscr F=(A,E,V)$.
    It is straightforward to show that $(\bullet^*,\bullet_*)$ is a translation
    from $\mathscr F$ to $\mathscr F$ where $\bullet^*:A\to A$ is the identify
    function on $A$ and $\bullet_*:E\to E$ is the identity function on $E$.
\end{proof}

\section{Comparing intelligence using ultrafilters and preserving
relative intelligence}
\label{ultrafiltersection}

The question of how to measure intelligence of RL agents is nontrivial,
even in a given fixed RL framework. One proposal is the Legg-Hutter intelligence
measure \cite{legg2007universal}, but that proposed measure involves
infinite sums and the noncomputable Kolmogorov complexity function, making the
proposal mathematically unwieldy; we have not been able to prove intelligence
preservation results in terms of Legg-Hutter intelligence. Instead, we will
compare intelligence using an approach which is mathematically more tractable,
originally introduced by \cite{alexander2019intelligence}.

The idea is that
in order to compare two RL agents $\pi$ and $\rho$,
to determine which one is more intelligent
(or whether they are equally intelligent), we can consider these three
possibilities to be
candidates in an election, where environments are voters. For any particular
environment $\mu$, if $V^\pi_\mu>V^\rho_\mu$, then $\mu$ votes that $\pi$ is
more intelligent than $\rho$. If $V^\pi_\mu<V^\rho_\mu$, then $\mu$ votes that
$\pi$ is less intelligent than $\rho$. If $V^\pi_\mu=V^\rho_\mu$, then $\mu$
votes that $\pi$ and $\rho$ are equally intelligent\footnote{Note that this
comparison method is based solely on binary comparisons of performance, ignoring
the actual numerical difference: that is, it does not matter whether
$V^\pi_\mu$ is $+1$ better or $+1000$ better than $V^\rho_\mu$, $\mu$ will
vote just as hard for $\pi$ over $\rho$ either way.}.
How
can we decide the winner of such an election?
It turns out there is an elegant way to do this using machinery from mathematical
logic known as \emph{ultrafilters} (we will motivate ultrafilters below, assuming
no prior knowledge thereof).

\subsection{Introduction to ultrafilters}
\label{ultrafiltermotivationsection}

We give an introduction to ultrafilters in terms of elections (they were previously
introduced this way in \cite{alexander2021measuring} and in
\cite{alexander2022big}). In this subsection, we fix a set $E$ of
environments. If the environments in $E$ vote in an election between finitely many
candidates, how can we determine which candidate wins?

Say that a subset $X\subseteq E$ is a \term{majority} if electoral victory would
already be guaranteed given only the votes of $X$. Can we think of any
axioms that majorities should satisfy?

Here are three fairly obvious axioms for majorities:
\begin{itemize}
    \item (Properness) $\emptyset$ is not a majority (if no-one votes for
    you, you lose).
    \item (Monotonicity) If $X$ is a majority and $Y\supseteq X$ then $Y$ is
    a majority (additional votes should do no harm).
    \item (Maximality) If $X$ is not a majority, then
    its complement $X^c$ is a majority (in
    a two-candidate election, if one candidate does not win, then the other
    candidate wins).
\end{itemize}

A fourth axiom is much less obvious, and in fact is highly counter-intuitive
if we rely on our intuition about \emph{finite-voter} elections. We would
probably never think of this next axiom if we were only thinking in terms of
elections, but recall that we are particularly interested in a special type of
election, namely, an intelligence-comparison election.
We would very much like for the resulting agent comparator to be
transitive. In other words,
consider RL agents $\pi,\rho,\sigma$. If the voters vote that $\pi$ is more
intelligent than $\rho$, and also they vote that $\rho$ is more intelligent
than $\sigma$, then we would very much desire that they should vote that
$\pi$ is more intelligent than $\sigma$. To say the voters vote $\pi$ more
intelligent than $\rho$ is to say that some majority $X$ votes as much, and
to say that they vote $\rho$ more intelligent than $\sigma$ is to say that
some majority $Y$ votes as much. Assuming \emph{individual} voters are
consistent, it would follow that $X\cap Y$ vote $\pi$ more intelligent than
$\sigma$. Thus, in order to achieve the desired transitivity, we enforce
the following counter-intuitive axiom.

\begin{itemize}
    \item ($\cap$-closure) If $X$ and $Y$ are majorities, then $X\cap Y$ is a
    majority.
\end{itemize}

It turns out that through these electoral considerations we have already arrived
at the mathematically sophisticated notion of the ultrafilter.

\begin{definition}
\label{ultrafilterdefn}
    Suppose $E$ is a set. By an \term{ultrafilter on $E$} we mean a
    set $\mathcal U$ of subsets of $E$ (intuitively thought of
    as \emph{majorities}) which satisfy Properness, Monotonicity,
    Maximality and $\cap$-closure.
\end{definition}

Thus, if the environments in the set $E$ are going to vote in an election
with finitely many candidates, one way to
determine the winner is to fix an ultrafilter $\mathcal U$ on $E$ and
declare that for each candidate $c$, if
$\{\mu\in E\,:\,\mbox{$\mu$ votes for $c$}\}\in\mathcal U$, then $c$ wins.
The $\cap$-closure and Properness axioms ensure
at most one candidate can win. The Maximality axiom (possibly iterated
if there are $>2$ candidates) ensures at least one candidate must win.
Economists have shown \cite{kirman} that if we impose certain
requirements on election-decision
methods, then conversely, every election-decision method satisfying those
requirements is one of these ultrafilter-based decision methods\footnote{The
requirements in question are exactly the desiderata from Arrow's Impossibility
Theorem, minus non-dictatorialness. Non-dictatorial decision-methods correspond
exactly with so-called \emph{free ultrafilters}: an ultrafilter $\mathcal U$ on $E$
is \term{free} if it has the property that there does not exist any $\mu\in E$
such that $\{\mu\}\in\mathcal U$. Assuming $E$ is infinite, it is known that
free ultrafilters on $E$ exist. This does not contradict Arrow's Impossibility
Theorem, because Arrow's Impossibility Theorem requires that the set of voters
is finite.}.

\subsection{Preservation of relative intelligence by transformation
of RL frameworks}

The previous subsection motivates the following notion of relative intelligence.

\begin{definition}
    Suppose $\mathscr F=(A,E,V)$ is an RL framework. Let $\mathcal U$ be an
    ultrafilter on $E$. We define the intelligence
    comparator $\leq_{\mathcal U}$, a binary relation on $A$, as follows.
    For all $\pi,\rho\in A$, we declare $\pi\leq_{\mathcal U}\rho$ iff
    $\{\mu\in E\,:\,V^\pi_\mu\leq V^\rho_\mu\}\in\mathcal U$.
\end{definition}

In plain English: $\pi\leq_{\mathcal U}\rho$ if the environments \emph{vote}
that $\rho$ performs at least as well as $\pi$ (when we use $\mathcal U$ to
decide the outcome of the election once the votes are cast).
We leave the proof of the following lemma as an exercise to the reader
(using the ultrafilter axioms, Definition \ref{ultrafilterdefn}).

\begin{lemma}
    Suppose $\mathscr F=(A,E,V)$ is an RL framework and $\mathcal U$ is
    an ultrafilter on $E$.
    \begin{enumerate}
        \item (Reflexivity) For every $\pi\in A$,
        $\pi\leq_{\mathcal U}\pi$.
        \item (Transitivity) For all $\pi,\rho,\sigma\in A$,
        if $\pi\leq_{\mathcal U}\rho$ and $\rho\leq_{\mathcal U}\sigma$,
        then $\pi\leq_{\mathcal U}\sigma$.
    \end{enumerate}
\end{lemma}

Since we are interested in preservation (or lack thereof) of relative intelligence 
by a transformation from one RL framework to another, we would like a way to
transform the above relative intelligence notion between frameworks.

\begin{definition}
\label{ultrafiltertranslationdefn}
    (Transformation of an ultrafilter)
    Suppose $\mathscr F=(A,E,V)$ and
    $\overline{\mathscr F}=(\overline A,\overline E,\overline V)$
    are RL frameworks,
    $(\pi\mapsto\pi^*:A\to \overline A,\mu\mapsto\mu_*:\overline E\to E)$
    is a transformation from $\mathscr F$
    to $\overline{\mathscr F}$, and $\mathcal U$ is an ultrafilter on
    $\overline E$.
    For each $Y\subseteq \overline E$,
    let $Y_*=\{\mu_*\,:\,\mu\in Y\}\subseteq E$.
    We define
    \[
        \mathcal U_*=\{X\subseteq E\,:\,
        \mbox{$X\supseteq Y_*$ for some $Y\in \mathcal U$}\}.
    \]
\end{definition}

\begin{lemma}
    For all $\mathscr F$,
    $\overline{\mathscr F}$,
    $(\bullet^*,\bullet_*)$,
    $\mathcal U$ as
    in Definition \ref{ultrafiltertranslationdefn},
    $\mathcal U_*$ is an ultrafilter on $E$.
\end{lemma}

\begin{proof}
    Straightforward.
\end{proof}

The following is a relative intelligence preservation theorem for RL framework
transformations, in the following sense. It says that if we have a transformation
from a source framework to a destination framework, and if we compare relative
intelligence in the destination framework using the electoral method (deciding
elections with some ultrafilter $\mathcal U$ on the destination framework's
environments), then those comparisons are
preserved by the transformation (if we decide elections with $\mathcal U_*$ on
the source framework's environments).

\begin{theorem}
\label{preservationthm}
    (Preservation Theorem)
    Suppose $\mathscr F=(A,E,V)$,
    $\overline{\mathscr F}=(\overline A,\overline E,\overline V)$
    are RL frameworks
    and $(\bullet^*:A\to \overline A,\bullet_*:\overline E\to E)$
    is a transformation from $\mathscr F$ to $\overline{\mathscr F}$.
    For any ultrafilter $\mathcal U$ on $\overline E$,
    the transformation $(\bullet^*,\bullet_*)$ preserves relative intelligence
    in the following sense:
    for all $\pi,\rho\in A$, we have $\pi\leq_{\mathcal U_*}\rho$
    iff $\pi^*\leq_{\mathcal U}\rho^*$.
\end{theorem}

\begin{proof}
    ($\Leftarrow$) Assume $\pi^*\leq_{\mathcal U}\rho^*$.
    By definition, this means $Y\in\mathcal U$, where
    $Y=\{\mu\in \overline E\,:\,\overline V^{\pi^*}_\mu
    \leq \overline V^{\rho^*}_\mu\}$.
    Let $X=\{\nu\in E\,:\,V^\pi_\nu\leq V^\rho_\nu\}$,
    we must show $X\in\mathcal U_*$.
    By the Monotonicity property of ultrafilters,
    it suffices to show some subset of $X$ is in $\mathcal U_*$.
    Since $Y\in\mathcal U$, it follows that $Y_*\in\mathcal U_*$;
    we will show $Y_*\subseteq X$. Compute:
    \begin{align*}
        Y_* &= \{\mu_*\,:\,\mu\in Y\}
                &\mbox{(Def.\ of $Y_*$)}\\
            &= \{\nu\in E\,:\,\mbox{$\nu=\mu_*$ for some $\mu\in Y$}\}
                &\mbox{(Rewriting)}\\
            &= \{\nu\in E\,:\,\mbox{$\nu=\mu_*$ for some $\mu\in \overline E$ with
                $\overline V^{\pi^*}_\mu\leq \overline V^{\rho^*}_\mu$}\}
                &\mbox{(Def.\ of $Y$)}\\
            &= \{\nu\in E\,:\,\mbox{$\nu=\mu_*$ for some $\mu\in \overline E$ with
                $V^\pi_{\mu_*}\leq V^\rho_{\mu_*}$}\}
                &\mbox{(Def.\ \ref{translationdefn} part 1)}\\
            &\subseteq \{\nu\in E\,:\,V^\pi_\nu\leq V^\rho_\nu\}\\
            &= X.
                &\mbox{(Def.\ of $X$)}
    \end{align*}

    ($\Rightarrow$) By rewriting our proof of $(\Leftarrow)$ with
    $\leq$ changed to $\nleq$ throughout, we get a proof that if
    $\pi^*\nleq_{\mathcal U} \rho^*$ then $\pi\nleq_{\mathcal U_*}\rho$.
\end{proof}

\begin{remark}
\label{measurementtheoryrmk}
    Readers interested in measurement theory will be interested in the following
    variation. Suppose $\mathscr F=(A,E,V)$,
    $\overline{\mathscr F}=(\overline A,\overline E,\overline V)$
    are RL frameworks and
    $(\bullet^*:A\to \overline A,\bullet_*:\overline E\to E)$
    is a transformation from $\mathscr F$ to $\overline{\mathscr F}$.
    Call $(\bullet^*,\bullet_*)$ a \term{scaling transformation} if it
    satisfies the following additional property:
    \begin{itemize}
        \item For all $\pi,\rho\in A$, for all $\mu\in \overline E$,
        for all $k\in\mathbb R$,
        $\overline V^{\pi^*}_\mu<k\overline V^{\rho^*}_\mu$
        iff $V^\pi_{\mu_*}<kV^\rho_{\mu_*}$.
    \end{itemize}
    Fix an ultrafilter $\mathcal U$ on $\overline E$.
    For all $\pi,\rho\in A$ and $k\in\mathbb R$,
    define:
    \begin{itemize}
        \item
        $\pi^*\leq_{\mathcal U}k\rho^*$
        iff $\{\mu\in \overline E
        \,:\,
        \overline V^{\pi^*}_\mu\leq k\overline V^{\rho^*}_\mu\}\in\mathcal U$;
        \item
        $\pi\leq_{\mathcal U_*}k\rho$
        iff $\{\nu\in E\,:\,V^\pi_\nu\leq kV^\rho_\nu\}\in\mathcal U_*$.
    \end{itemize}
    Then by almost identical reasoning to the proof of
    Theorem \ref{preservationthm},
    one can show that for any scaling transformation
    $(\bullet^*,\bullet_*)$, $\pi\leq_{\mathcal U_*}k\rho$
    iff $\pi^*\leq_{\mathcal U}k\rho^*$.
    Thus, scaling transformations preserve relative intelligence even more strongly:
    they preserve real ratio relations such as ``$\pi$ is at least twice
    as intelligent as $\rho$'' or ``$\pi$ is not at least half as intelligent
    as $\rho$''.
\end{remark}

\section{Concrete results}

In this section we will introduce four specific concrete RL frameworks.
This will involve fixing action-sets and percept-sets, defining histories,
and defining specific performance measures $V^\pi_\mu$. Bear in mind that
this infrastructure is specific to the four specific concrete RL frameworks.

Fix nonempty finite sets $\mathcal A$, $\mathcal E$.
Elements of $\mathcal A$ are \term{actions} and elements of $\mathcal E$ are
\term{percepts}.
We assume $\mathcal A\cap \mathcal E=\emptyset$.
We typically write a member of $\mathcal A$ as $x$ and a member $\mathcal E$ as $y$.
Assume a function $R:\mathcal E\to\mathbb Z$ assigning\footnote{By abstracting
the function $R$ out, rather than requiring
that percepts be observation-reward pairs, we simplify certain technical details.
A similar device is used in \cite{feys2018long}. See
also \cite{legg2013approximation} where two different versions of RL are
implemented, one with observation-reward pairs, one with reward-observation
pairs, in order to test whether some empirical results depend on the ordering
of the pairs.} to
every percept $x\in \mathcal E$ an integer-valued \term{reward}.
We assume the range of $R$ includes $0$ and $1$.

The fact that $R(x)$ is integer-valued is critical for some of the proofs
below. We do not currently know whether the theorems in question would remain
true if $R(x)$ were allowed to be an arbitrary element of $\mathbb Q$,
which would be more relevant in practice: rewards in reinforcement learning
are not usually restricted to be integer-valued (though some important environments
do have integer-valued rewards, for example environments where the agent gets
reward $+1$ for winning a game, $-1$ for losing a game, and $0$ for any other
move).

We define \term{histories} inductively so that:
(i) The empty sequence is a history;
(ii) for any percept $x$, $\langle x\rangle$ is a history;
(iii) for any nonempty history $h$ ending with a percept,
for any action $y$, $hy$ is a history (where $hy$ is the result
of appending $y$ to $h$);
(iv) for any nonempty history $h$ ending with an action,
for any percept $x$, $hx$ is a history (where $hx$ is the result
of appending $x$ to $h$).
In plain English: a history is a finite sequence starting with
a percept, followed by an action, followed by a percent, followed
by an action, and so on for some finite number of steps
(the empty sequence is also considered a history).
An \term{agent history} is a history which ends with a percept
(so named because these are the histories intended to be seen by
the agent). An \term{environment history} is a history which is
either empty or ends with an action. We write $H_A$ for the set of all
agent histories, and we write $H_E$ for the set of all environment
histories.

\begin{lemma}
    A history is an agent history iff it has odd length; it is an environment
    history iff it has even length.
\end{lemma}

\begin{proof}
    By induction.
\end{proof}

For nonempty finite set $X$, let $\Delta(X)$ be the set of $\mathbb Q$-valued probability distributions on $X$.

\begin{definition}[Deterministic and Stochastic Agents and Environments] Define the following sets:
    \begin{itemize}
        \item[]
         $A^{\det}=\mathcal{A}^{H_{A}}$, the set of all functions
        $\pi:H_A\to\mathcal A$ (call these functions
        \term{deterministic agents}).
        \item[]
    $E^{\det}=\mathcal{E}^{H_E}$, the set of all functions
        $\mu:H_E\to\mathcal E$ (call these functions
        \term{deterministic environments}).
        \item[]
      $A^{\nondet}=\Delta(\mathcal A)^{H_A}$, the set of
      all functions $\pi:H_A\to\Delta(\mathcal A)$
        (call these functions \term{stochastic agents}).
        \item[]
    $E^{\nondet}=\Delta(\mathcal E)^{H_E}$, the set of all
        functions $\mu:H_E\to\Delta(\mathcal E)$
        (call these functions \term{stochastic environments}).
    \end{itemize}
\end{definition}

An environment provides a way to obtain a percept from an environment history.
Appending that percept to that history yields an agent history. An agent then
provides a way to obtain an action from an agent history, and appending that
action to that agent history gives us another environment history, and the
process can be repeated forever.

\begin{definition}[Expected total reward]
    Let $\pi\in A^{\det}\cup A^{\nondet}$, $\mu\in E^{\det}\cup E^{\nondet}$.
    \begin{enumerate}
        \item
        For every $n\in\mathbb N$, let $V^\pi_{\mu,n}$ be the
        expected value of the total reward-sum $R(x_1)+\cdots+R(x_n)$
        if the history $x_1y_1\ldots x_ny_n$ is generated as
        follows:
        \begin{itemize}
            \item If $\mu\in E^{\det}$ then $x_1=\mu(\langle\rangle)$.
            If $\mu\in E^{\nondet}$ then $x_1$ is randomly chosen from
            $\mathcal E$ based on the probability distribution
            $\mu(\langle\rangle)\in\Delta(\mathcal E)$.
            \item If $\pi\in A^{\det}$ then $y_1=\pi(\langle x_1\rangle)$.
            If $\pi\in A^{\nondet}$ then $y_1$ is randomly chosen from
            $\mathcal A$ based on the probability distribution
            $\pi(\langle x_1\rangle)\in\Delta(\mathcal A)$.
            \item For $1<i<n$, if $\mu\in E^{\det}$ then
                $x_{i+1}=\mu(x_1y_1\ldots x_iy_i)$; if $\mu\in E^{\nondet}$
                then $x_{i+1}$ is randomly chosen from $\mathcal E$ based
                on the probability distribution
                $\mu(x_1y_1\ldots x_iy_i)\in \Delta(\mathcal E)$.
            \item For $1<i\leq n$, if $\pi\in A^{\det}$ then
                $y_i=\pi(x_1y_1\ldots x_{i-1}y_{i-1}x_i)$; if
                $\pi\in A^{\nondet}$ then $y_i$ is randomly chosen from
                $\mathcal A$ based on the probability distribution
                $\pi(x_1y_1\ldots x_{i-1}y_{i-1}x_i)$.
        \end{itemize}
        \item
        Let $V^\pi_\mu=\lim_{n\to\infty}V^\pi_{\mu,n}$,
        provided the limit converges to a real number. If not, then
        $V^\pi_\mu$ is undefined.
    \end{enumerate}
\end{definition}

Since the above $V^\pi_\mu$ does not always converge, it is not directly
suitable for Definition \ref{rlframeworkdefn}. To get around this, we
restrict our attention to environments $\mu$ for which $V^\pi_\mu$ always
converges (this trick was introduced in \cite{alexanderhutter}).

\begin{definition}[Well-behaved environments]
\,
    \begin{enumerate}
        \item
        We say
        $\mu\in E^{\det}\cup E^{\nondet}$ is \term{well-behaved} if it has the
        following property:
        for every $\pi\in A^{\det}\cup A^{\nondet}$,
        $V^\pi_\mu$ exists.
        \item
        Let $W^{\det}$ denote the set of well-behaved deterministic environments
        and $W^{\nondet}$ the set of well-behaved stochastic
        environments.
    \end{enumerate}
\end{definition}

A word on notation might be helpful.
In the notation $V^\pi_\mu$, the superscript on $V$ is used for
the agent, and the subscript on $V$ is used for the environment.
In the same way, in the following definition, the superscript on
$\mathscr F$ refers to agents, and the subscript on $\mathscr F$
refers to environments.

\begin{definition}[Four Specific RL Frameworks]
\label{fourspecificframeworksdef}
\,
    \begin{itemize}
        \item
        The \term{standard RL framework with deterministic
        agents and environments} is the RL framework
        $\mathscr F^{\det}_{\det}=(A^{\det},W^{\det},V)$.
        \item
        The \term{standard RL framework with stochastic agents and
        deterministic environments} is the RL framework
        $\mathscr F^{\nondet}_{\det}=(A^{\nondet},W^{\det},V)$.
        \item
        The \term{standard RL framework with deterministic
        agents and stochastic environments} is the RL framework
        $\mathscr F^{\det}_{\nondet}=(A^{\det},W^{\nondet},V)$.
        \item
        The \term{standard RL framework with stochastic
        agents and environments} is the RL framework
        $\mathscr F^{\nondet}_{\nondet}=(A^{\nondet},W^{\nondet},V)$.
    \end{itemize}
\end{definition}

The following theorem is the main result of this paper.
For the four concrete frameworks of Definition \ref{fourspecificframeworksdef},
we answer the $4\cdot (4-1)=12$ transformation-existence questions.

\begin{figure}
    \begin{center}
        \includegraphics[scale=1.0]{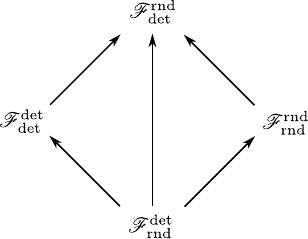}
    \end{center}
    \caption{Existence of transformations between four concrete RL
    frameworks.}
    \label{diamondfigure}
\end{figure}

\begin{theorem}
\label{mainthm}
    For all
    $
        \mathscr G,\mathscr H
        \in
        \{
            \mathscr F^{\nondet}_{\nondet},
            \mathscr F^{\nondet}_{\det},
            \mathscr F^{\det}_{\nondet},
            \mathscr F^{\det}_{\det}
        \}
    $
    with $\mathscr G\not=\mathscr H$, there is a transformation from
    $\mathscr G$ to $\mathscr H$ iff $\mathscr G=\mathscr F^{\nondet}_{\det}$
    or $\mathscr H=\mathscr F^{\det}_{\nondet}$. In other words:
    there is a transformation from $\mathscr G$ to $\mathscr H$
    if and only if there
    is an arrow from $\mathscr G$ to $\mathscr H$ in Figure \ref{diamondfigure}.
\end{theorem}

We will prove Theorem \ref{mainthm} by a series of prelimary results.

\subsection{Proof of Theorem \ref{mainthm}}

First, we will prove the positive parts of Theorem \ref{mainthm}.
We begin by defining embeddings of deterministic agents (resp.\ environments)
among stochastic agents (resp.\ environments).
Intuitively, these embeddings explain the positive parts of Theorem \ref{mainthm}.

\begin{lemma}
\label{deterministicasnondeterministiclemma}
    \begin{enumerate}
        \item
        There exists a function $\hat{\bullet}:A^{\det}\to A^{\nondet}$
        such that for all $\mu\in W^{\det}\cup W^{\nondet}$,
        $V^\pi_\mu=V^{\hat\pi}_\mu$.
        \item
        There exists a function $\hat{\bullet}:W^{\det}\to W^{\nondet}$
        such that for all $\pi\in A^{\det}\cup A^{\nondet}$,
        $V^\pi_\mu=V^\pi_{\hat\mu}$.
    \end{enumerate}
\end{lemma}

\begin{proof}
    \item
    (1)
    Define $\hat{\pi}:H_A\to\Delta(\mathcal A)$ by
    \[
        \hat{\pi}(y|h)
        =
        \begin{cases}
            1 & \mbox{if $y=\pi(h)$,}\\
            0 & \mbox{otherwise.}
        \end{cases}
    \]
    For any $\mu\in W^{\det}\cup W^{\nondet}$,
    by induction on $n$, it is easy to show that for each $n$,
    $V^{\hat\pi}_{\mu,n}=V^\pi_{\mu,n}$.
    Taking the limit as $n\to\infty$, we are done.

    \item
    (2) Similar to (1), defining $\hat{\mu}:H_E\to\Delta(\mathcal E)$
    by
    \[
        \hat{\mu}(x|h)
        =
        \begin{cases}
            1 & \mbox{if $x=\mu(h)$,}\\
            0 & \mbox{otherwise.}
        \end{cases}
    \]
\end{proof}

\begin{proposition}[Positive parts of Theorem \ref{mainthm}]
\label{positivethm}
For every arrow in Figure \ref{diamondfigure} from a source RL framework
to a destination RL framework, there is a transformation from said source
framework to said destination framework.
\end{proposition}

\begin{proof}
    (From $\mathscr F^{\det}_{\det}$ to $\mathscr F^{\nondet}_{\det}$)
    Define $\bullet^*:A^{\det}\to A^{\nondet}$ by $\pi^*=\hat\pi$ and
    define $\bullet_*:W^{\det}\to W^{\det}$ by $\mu_*=\mu$,
    where $\hat\pi$ is as in 
    Lemma \ref{deterministicasnondeterministiclemma}.
    It is straightforward to show $(\bullet^*,\bullet_*)$ is a transformation
    from $\mathscr F^{\det}_{\det}$ to $\mathscr F^{\nondet}_{\det}$
    (for Nontriviality 1 and Nontriviality 2 (Definition \ref{translationdefn}),
    use the fact the range of $R$ includes $0$ and $1$).

    \item
    (From $\mathscr F^{\det}_{\nondet}$ to $\mathscr F^{\det}_{\det}$)
    Define $\bullet^*:A^{\det}\to A^{\det}$ by $\pi^*=\pi$
    and define $\bullet_*:W^{\nondet}\to W_{\det}$ by $\mu_*=\hat\mu$,
    where $\hat\mu$ is as in
    Lemma \ref{deterministicasnondeterministiclemma}. It
    is straightforward to show $(\bullet^*,\bullet_*)$ is a transformation
    from $\mathscr F^{\det}_{\nondet}$ to $\mathscr F^{\det}_{\det}$.


    \item
    The other three arrows are similar.
\end{proof}

To prove the negative parts of Theorem \ref{mainthm}, we will need to
take mixtures of agents and environments.

\begin{lemma}[Mixing Lemma]
\label{mixingthm}

    \begin{enumerate}
        \item
        Given weights $w_1,\ldots,w_n\in (0,1)\cap \mathbb Q$,
        with $w_1+\cdots+w_n=1$,
        and agents $\pi_1,\ldots,\pi_n\in A^{\nondet}$,
        there exists $\pi\in A^{\nondet}$
        such that for every $\mu\in W^{\det}\cup W^{\nondet}$,
        $V^\pi_\mu = w_1V^{\pi_1}_\mu+\cdots+w_nV^{\pi_n}_\mu$.
        \item
        Given weights
        $w_1,\ldots,w_n\in(0,1)\cap\mathbb Q$, with $w_1+\cdots+w_n=1$,
        and environments $\mu_1,\ldots,\mu_n\in W^{\nondet}$,
        there exists $\mu\in W^{\nondet}$
        such that for all $\pi\in A^{\det}\cup A^{\nondet}$,
        $V^\pi_\mu=w_1V^\pi_{\mu_1}+\cdots+w_nV^\pi_{\mu_n}$.
    \end{enumerate}
\end{lemma}

\begin{proof}
    (1) For every history $h$ and every $\rho\in A^{\nondet}$,
    let $P^\rho(h)$ be the
    probability that $h$ would be an initial segment of the
    percept-action sequence that would be randomly generated if $\rho$
    interacted with some environment, subject to the condition that that
    environment initially outputs the percepts in $h$.
    As in \cite{aistats}, define $\pi:H_A\to\Delta(\mathcal A)$ by
    \[
        \pi(y|h) = \frac{
            w_1P^{\pi_1}(hy) + \cdots + w_nP^{\pi_n}(hy)
        }{w_1P^{\pi_1}(h) + \cdots + w_nP^{\pi_n}(h)}
    \]
    provided the denominator is $\not=0$, or $\pi(y|h)=1/|\mathcal A|$
    otherwise. By an inductive argument on $k$, the expected total reward
    $V^\pi_{\mu,k}$ which $\pi$ would obtain after $k$ steps interacting with
    any well-behaved environment $\mu$ equals
    $w_1V^{\pi_1}_{\mu,k}+\cdots+w_nV^{\pi_n}_{\mu,k}$, the weighted average
    of the expected rewards $\pi_1,\ldots,\pi_n$ would obtain after $k$ steps
    interacting with $\mu$. Taking the limit as $k\to\infty$,
    we are done. For details, see \cite{aistats}.

    (2) Similar to (1), with $\mu:H_E\to\Delta(\mathcal E)$ defined as follows.
    For every history $h$ and every $\nu\in W^{\nondet}$, let $P_\nu(h)$
    be the probability that $h$ would be an initial segment of the percept-action
    sequence that would be randomly generated if $\nu$ interacted with some agent,
    subject to the condition that that agent initially outputs the actions in
    $h$. Define
    \[
        \mu(x|h) = \frac{
            w_1P_{\mu_1}(hx) + \cdots + w_nP_{\mu_n}(hx)
        }{w_1P_{\mu_1}(h) + \cdots + w_nP_{\mu_n}(h)}
    \]
    provided the denominator is $\not=0$, or $\mu(x|h)=1/|\mathcal E|$
    otherwise. For details, adapt the proof of Lemma 48 (part 4) of
    \cite{aistats} to a finite vector of weights (the ``strongly well-behaved''
    hypothesis of said lemma can be replaced by ``well-behaved'' because of the
    finiteness of the vector of weights).
\end{proof}

\begin{theorem}
\label{firstimpossibilityresult}
    There does not exist a transformation from
    $\mathscr F^{\det}_{\det}$ to $\mathscr F^{\nondet}_{\nondet}$.
\end{theorem}

\begin{proof}
    For sake of contradiction, assume
    $(\bullet^*:A^{\det}\to A^{\nondet},\bullet_*:W^{\nondet}\to W^{\det})$ is
    a transformation.
    By Nontriviality 2 (Definition \ref{translationdefn}), pick
    $\pi\in A^{\det}$ and $\mu,\nu\in W^{\nondet}$ such that
    $V^{\pi^*}_\mu<V^{\pi^*}_\nu$.
    For every $\alpha\in (0,1)\cap \mathbb Q$, using
    Theorem \ref{mixingthm} (part 2)
    (with $n=2$, $w_1=\alpha$, $w_2=1-\alpha$), let $\sigma_\alpha$ be the
    result of mixing $\mu$ and $\nu$, giving $\alpha$ weight to $\mu$
    and $1-\alpha$ weight to $\nu$, so
    $V^{\pi^*}_{\sigma_\alpha}=\alpha V^{\pi^*}_{\mu} + (1-\alpha)V^{\pi^*}_{\nu}$.
    Thus for all such rational $\alpha<\beta$,
    we have $V^\pi_{(\sigma_{\alpha})_*}<V^\pi_{(\sigma_{\beta})_*}$.
    Choose rationals $\ell_1<\ell_2<\ell_3<\cdots$ in $(0,\frac12)\cap\mathbb Q$
    and choose
    rational $r\in(\frac12,1)\cap\mathbb Q$, so $r>\ell_i$ for every $i$.
    For each $i$, let $\tau_i=\sigma_{\ell_i}$.
    We have $V^\pi_{(\tau_1)_*}<V^\pi_{(\tau_2)_*}<\cdots$,
    and yet $V^\pi_{(\sigma_r)_*}>V^\pi_{(\tau_i)_*}$ for every $i$.
    This is impossible since $V^\pi_{(\sigma_r)_*}\in\mathbb Z$
    and each $V^\pi_{(\tau_i)_*}\in\mathbb Z$:
    there does not exist an infinite strictly ascending sequence of integers
    \emph{and} another integer bigger than them all.
\end{proof}

\begin{theorem}
\label{secondimpossibilityresult}
    There does not exist a transformation from
    $\mathscr F^{\nondet}_{\nondet}$ to
    $\mathscr F^{\det}_{\det}$.
\end{theorem}

\begin{proof}
    For sake of contradiction, assume
    $(\bullet^*:A^{\nondet}\to A^{\det},\bullet_*:W^{\det}\to W^{\nondet})$
    is a transformation. By Nontriviality 1 (Definition \ref{translationdefn}),
    pick $\pi,\rho\in A^{\nondet}$ and $\mu\in W^{\det}$
    such that $V^{\pi^*}_\mu<V^{\rho^*}_\mu$,
    so $V^\pi_{\mu_*}<V^\rho_{\mu_*}$.
    Using Theorem \ref{mixingthm} (part 1)
    (with $n=2$, $w_1=\alpha$, $w_2=1-\alpha$), for every
    $\alpha\in(0,1)\cap\mathbb Q$, let $\sigma_\alpha$ be
    the result of mixing $\pi$ and $\rho$, giving $\alpha$
    weight to $\pi$ and $1-\alpha$ weight to $\rho$.
    So $V^{\sigma_\alpha}_{\mu_*}
    =\alpha V^\pi_{\mu_*}+(1-\alpha)V^\rho_{\mu_*}$.
    Thus for all such rationals $\alpha<\beta$,
    we have $V^{\sigma_{\alpha}}_{\mu_*}
    <V^{\sigma_{\beta}}_{\mu_*}$,
    so $V^{\sigma_{\alpha}^*}_\mu
    < V^{\sigma_{\beta}^*}_\mu$.
    Choose $\ell_1<\ell_2<\cdots$ in $(0,\frac12)\cap\mathbb Q$
    and $r\in (\frac12,1)\cap\mathbb Q$, so $r>\ell_i$ for each $i$.
    For every $i$, let $\tau_i=\sigma_{\ell_i}$.
    We have $V^{\tau_1^*}_\mu
    <V^{\tau_2^*}_\mu<\cdots$,
    and yet $V^{\sigma_r^*}_\mu>V^{\tau_i^*}_\mu$
    for each $i$. This is impossible for the same reason as
    in Theorem \ref{firstimpossibilityresult}.
\end{proof}

\begin{theorem}
\label{negativethm}
    (Negative parts of Theorem \ref{mainthm})
    For all distinct RL frameworks $\mathscr G$ and $\mathscr H$ in
    Figure \ref{diamondfigure}, if the figure does not include an arrow
    from $\mathscr G$ to $\mathscr H$, then there is no transformation from
    $\mathscr G$ to $\mathscr H$.
\end{theorem}

\begin{proof}
    (From $\mathscr F^{\det}_{\det}$ to $\mathscr F^{\nondet}_{\nondet}$
    and from $\mathscr F^{\nondet}_{\nondet}$ to $\mathscr F^{\det}_{\det}$)
    By Theorems \ref{firstimpossibilityresult}
        and \ref{secondimpossibilityresult}.

    \item
    (From $\mathscr F^{\nondet}_{\det}$ to $\mathscr F^{\det}_{\det}$)
    By Theorem \ref{positivethm},
    there is a transformation from $\mathscr F^{\nondet}_{\nondet}$
    to $\mathscr F^{\nondet}_{\det}$.
    If there were a transformation from $\mathscr F^{\nondet}_{\det}$
    to $\mathscr F^{\det}_{\det}$, then by
    composability (Lemma \ref{transitivetranslationlemma}),
    there would be a transformation from
    $\mathscr F^{\nondet}_{\nondet}$ to $\mathscr F^{\det}_{\det}$.
    This would violate the previous case.

    \item
    (From $\mathscr F^{\det}_{\det}$ to $\mathscr F^{\det}_{\nondet}$)
    By Theorem \ref{positivethm},
    there is a transformation from $\mathscr F^{\det}_{\nondet}$ to
    $\mathscr F^{\nondet}_{\nondet}$.
    If there were a transformation from $\mathscr F^{\det}_{\det}$
    to $\mathscr F^{\det}_{\nondet}$, then by composability
    (Lemma \ref{transitivetranslationlemma}),
    there would be a transformation from
    $\mathscr F^{\det}_{\det}$ to $\mathscr F^{\nondet}_{\nondet}$.
    This would violate the first case.

    \item
    The remaining three negative results are proved similarly.
\end{proof}

Combining Proposition \ref{positivethm} and Theorem \ref{negativethm},
we have proved Theorem \ref{mainthm}.

Theorem \ref{mainthm} suggests that, at least if rewards are limited to
integers, the nature of reinforcement learning may be inherently
different depending whether agents be deterministic or stochastic,
and whether environments be deterministic or stochastic. We do not currently
know whether Theorem \ref{mainthm} would remain true if arbitrary rational-number
rewards were allowed. For lack of any better evidence, though, the analysis here
at least urges that researchers should exercise caution before speaking about
RL as if these decisions don't matter.

The positive parts of Theorem \ref{mainthm} can be slightly strengthened:
it can be shown that the transformations in the proof of Theorem \ref{positivethm}
are in fact scaling transformations,
in the sense of Remark \ref{measurementtheoryrmk}.

\section{Summary and conclusion}

This paper is intended as a tentative initial step toward the difficult problem
of comparing different reinforcement learning frameworks in general.
Different authors all have the same high-level intuition about RL, but the
formal details vary from author to author: which formalizations of RL are
equivalent to each other, and which formalizations are fundamentally
different? As an extreme example, a version of RL where rewards
are required to always be $0$, is \emph{clearly} weaker than all ordinary
versions of RL. But what does that formally even mean?

We introduced (Definition \ref{translationdefn}) the notion of
a \emph{transformation} from one RL framework to another.
In Section \ref{ultrafiltersection} we recalled
from \cite{alexander2019intelligence} an elegant method of comparing RL agent
intelligence based on electoral considerations. The high-level idea is to
consider RL environments to be voters who vote (based on the performance of
agents in those environments) to decide whether one agent is
more intelligent than another (or whether both are equally intelligent).
These intelligence-competition elections have to be
decided somehow, and economists in the 1970s showed that the election decision
procedures satisfying certain desiderata correspond exactly to so-called
\emph{ultrafilters}, which we recalled for the reader who might not be
familiar with them (Subsection \ref{ultrafiltermotivationsection}).
We showed (Theorem \ref{preservationthm}) that if a transformation exists from a
source RL framework to a destination RL framework, then this transformation can be
used to transform ultrafilter-based intelligence comparators in the destination
framework into ultrafilter-based intelligence comparators in the source
framework, in such a way as to preserve the relative intelligence of agents.

The reason for introducing transformations is that we intend them to be a proxy for
the intuitive notion of one RL framework being reducible to another. Given two
RL frameworks, if each is reducible to the other in this sense, then that serves
as a proxy for the intuitive notion of the two frameworks being equivalent.

We introduce (Definition \ref{fourspecificframeworksdef}) four concrete RL
frameworks, differing from each other only in terms of two binary parameters:
(i) whether agents are deterministic or stochastic, and (ii) whether
environments are deterministic or stochastic. This gives rise to
$4\cdot 3=12$ questions about existence of transformations. We answer all twelve
questions, five positively and seven negatively (Theorem \ref{mainthm}); no
two of the four frameworks are equivalent in the sense of having transformations
going in both directions. This is evidence suggesting that all four
frameworks might be mutually non-equivalent.

Our high-level hope is that these results will encourage authors to be more
specific, when talking about reinforcement learning, about which version of RL
they mean. For example, when Silver et al suggest \cite{silver2021reward}
that RL will lead to AGI, which version of RL do they mean? (Silver et al do
claim to provide a definition in that paper, but their definition is not
rigorous, for example it is unclear exactly which numbers rewards are allowed to
be.) Furthermore, we hope that the question of existence of transformations
from one RL framework to another will be a source of much interesting mathematics.

Our RL framework definition is quite general but not as general as possible.
More extreme variations of RL will require, in future work, more general RL
framework notions 
(but the ideas in this paper serve as a template for how one can explore
intelligence preservation results in those more general frameworks)\footnote{
Some such variations
include, for example, (i)
RL with multiple reward-signals, (ii) RL with multiple agents
\cite{littman1996generalizedx,beggs2005convergence,collective-1993,multiagent-1994-most-cited,multiagent-2021-most-cited,hernandez2011more}, (iii) preference-based RL \cite{wirth2017survey}, (iv) RL with rewards from non-Archimedean number systems
        allowing infinitary or infinitesimal rewards
        (suggested by \cite{alexander2020archimedean}, implicitly suggested
        by \cite{zhao2009reinforcement}, and conspicuously not ruled out
        by \cite{silver2021reward}), (v) along the same lines, RL involving
        non-Archimedean probabilities \cite{pedersen2014comparative}, (vi) RL where $V^\pi_\mu$ is allowed to diverge (and relative performance of
        agents in an environment is defined in various ways accordingly).
}.

\section*{Acknowledgments}

We acknowledge Cole Wyeth and Aram Ebtakar for comments and feedback.

\bibliographystyle{plain}
\bibliography{main}
\end{document}